\documentclass[11pt,letterpaper]{article}
\usepackage[letterpaper]{geometry}
\usepackage{acl2012}
\usepackage{times}
\usepackage{latexsym}
\usepackage{multirow}
\usepackage{url}
\usepackage{graphicx}
\usepackage{xcolor}
\usepackage{amsmath, amssymb}
\usepackage{algorithm,algpseudocode}
\usepackage{algorithmicx}
\usepackage{longtable}
\usepackage{tikz}
\usepackage{pgfplots}
\usepackage{pgf-umlsd}
\usepackage{ifthen}

\makeatletter
\newcommand{\@BIBLABEL}{\@emptybiblabel}
\newcommand{\@emptybiblabel}[1]{}
\makeatother
\usepackage[hidelinks]{hyperref}

\title{A new dataset and model for \\ learning to understand navigational instructions}

\author{Ozan Arkan Can \\
  Computer Science and Engineering \\
  Ko\c{c} University \\
  Istanbul, Turkey\\
  {\tt ocan13@ku.edu.tr} \\\And
  Deniz Yuret \\
  Computer Science and Engineering \\
  Ko\c{c} University \\
  Istanbul, Turkey\\
  {\tt dyuret@ku.edu.tr} \\}

\date{}

\begin{document}
\maketitle
\begin{abstract}
In this paper, we present a state-of-the-art model and introduce a new dataset for grounded language learning.  Our goal is to develop a model that can learn to follow new instructions given prior instruction-perception-action examples.  We based our work on the SAIL dataset which consists of navigational instructions and actions in a maze-like environment.  The new model we propose achieves the best results to date on the SAIL dataset by using an improved perceptual component that can represent relative positions of objects.  We also analyze the problems with the SAIL dataset regarding its size and balance.  We argue that performance on a small, fixed-size dataset is no longer a good measure to differentiate state-of-the-art models. We introduce SAILx, a synthetic dataset generator, and perform experiments where the size and balance of the dataset are controlled.
\end{abstract}

\section{Introduction}

% Most of natural language processing tasks (parsing, machine translation, tagging, semantic role labeling etc.) deal with the language only in text form. In these tasks, semantics is usually represented with symbolic logic or other words \cite{miller1995wordnet,saint1995computational,liu2004conceptnet}. However, a child acquires her mother language in a different way, she connects words with objects and actions by processing environmental observations. The grounded language learning studies aim to capture the meaning by linking the language with the physical world that it is used in \cite{harnad1990symbol}. 

This paper explores the task of learning to follow natural language instructions in the simple domain of navigating a maze-like environment.  We propose a new model with a unique perceptual representation and use the SAIL dataset \cite{macmahon2006walk} to compare our model with previous work. We describe the specifics of the dataset and the problems with its size and balance below. To address these problems we introduce a synthetic data generator SAILx and perform experiments where we control the size and balance of the datasets.

\begin{figure}[h]
	\label{task}
    \centering
    \def\svgwidth{\columnwidth}
    \resizebox{0.85\columnwidth}{0.21\textheight}{
    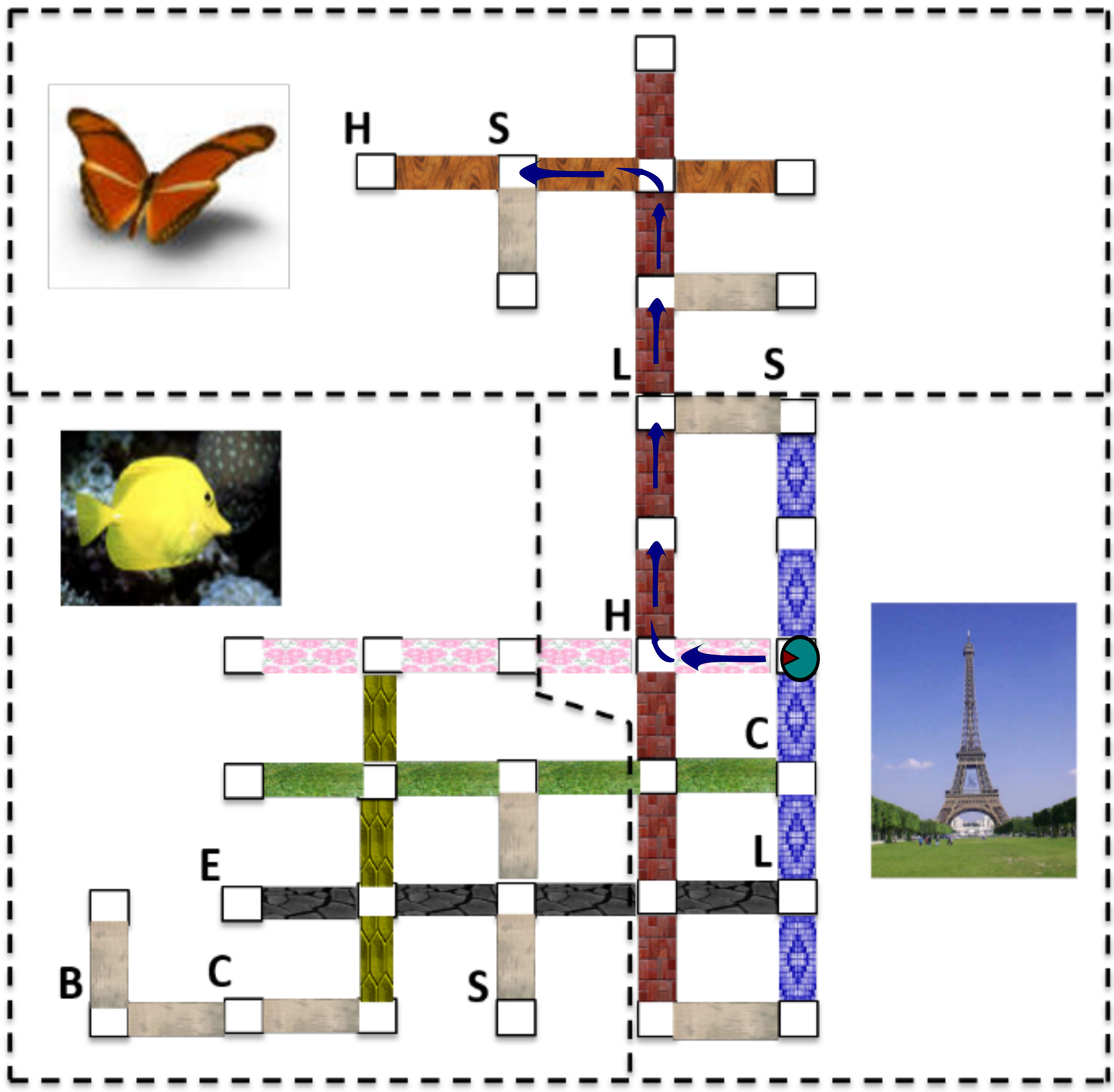
    }
    \caption{An illustration of a map and a set of instructions from the SAIL dataset \protect \cite{macmahon2006walk}. The letters indicate items (e.g. S for Sofa), the figures indicate wall paintings for each area divided by dashed lines, and the floor patterns distinguish the flooring. The circle represents the initial position of the agent and blue arrows represent the execution of the instruction set: ``Take the pink path to the red brick intersection. Go right on red brick. Go all the way to the wood intersection. Go left on wood. Position one is where the sofa is.''}
\end{figure}

\begin{table*}[ht]
\centering
\resizebox{\textwidth}{!}{ 
\begin{tabular}{lll}
\multicolumn{1}{c}{Task and frequency} & \multicolumn{1}{c}{Definition}                                                                                                                                                                                                                              & \multicolumn{1}{c}{Examples}                                                                     \\ \hline
Language only (31.7\%)           & \begin{tabular}[c]{@{}l@{}}Instructions contain only linguistic dependencies.\\ The perceptual information is not required to solve the task.\end{tabular}                                                                                                  & \begin{tabular}[c]{@{}l@{}}turn right, move two steps,\\ turn right and take a step\end{tabular} \\ \hline
Turn to X (7.0\%)               & \begin{tabular}[c]{@{}l@{}}The agent should understand the specified perceptual phenomenon\\ and take one or two "turn" actions.\end{tabular}                                                                                                               & \begin{tabular}[c]{@{}l@{}}turn to the chair,\\ turn to the green path\end{tabular}              \\ \hline
Move to X (13.4\%)               & \begin{tabular}[c]{@{}l@{}}The agent should understand the specified perceptual phenomenon\\ and take one or more "move" actions.\end{tabular}                                                                                                              & \begin{tabular}[c]{@{}l@{}}move to the sofa,\\ go to the end\end{tabular}                        \\ \hline
\begin{tabular}[c]{@{}l@{}}Turn and\\ 
Move to X (1.7\%)\end{tabular}       & \begin{tabular}[c]{@{}l@{}}The agent should understand the specified perceptual phenomenon.\\ The agent should first complete orientation step(s), then the movement(s).\end{tabular}                                                                       & \begin{tabular}[c]{@{}l@{}}turn and move to the chair,\\ go to the easel\end{tabular}            \\ \hline
Orient (5.2\%)                  & \begin{tabular}[c]{@{}l@{}}The agent should orient itself along the specified perceptual phenomenon.\\ There might be more than one conditions.\end{tabular}                                                                                                & turn so that the wall is on your back                                                            \\ \hline
Description (9.6\%)                & \begin{tabular}[c]{@{}l@{}}Instructions describe a specific agent-environment configuration. \\ This type of instructions are generally used to describe the final position.\\ The most of the instructions require to take only STOP action.\end{tabular} & \begin{tabular}[c]{@{}l@{}}you should be at the intersection of\\ blue and brown\end{tabular} \\ \hline
Move Until (8.7\%)              & \begin{tabular}[c]{@{}l@{}} The agent should move in the forward direction until a specific perceptual\\
phenomenon occurs.\end{tabular}                                                                                                                                                                     & walk forward until you reach the blue floors                                                     \\ \hline
Any combination (22.7\%)         & \begin{tabular}[c]{@{}l@{}}Instructions contain two parts and at least one of them\\ has a perceptual dependency.\end{tabular}                             & at the black road intersection take a left                                                       \\ \hline
\end{tabular}}
\caption{Task definitions and examples.}
\label{taskdef}
\end{table*}

In the SAIL dataset (Figure~\ref{task}, Section~\ref{saildataset}), an agent in a maze like environment receives sensory information from its line of sight. The agent is asked to navigate from a starting position to a target position which is described by a free-form natural language instruction. The aim of the agent is to follow the instruction by generating a sequence of actions. The agent can take one of four possible actions, \{\textit{MOVE, RIGHT, LEFT, STOP}\}. The \textit{RIGHT} and \textit{LEFT} actions change the orientation of the agent, where the \textit{MOVE} action transports the agent to the next position in the direction it is facing. The agent ends its trip if it takes the \textit{STOP} action, hits a wall, or exceeds the maximum number of actions.

Joint modeling of language and the world with descriptive representations is a crucial requirement for grounded language acquisition. % solving the navigational instruction following problem. 
Previous studies \cite{chen2011learning,chen2012fast,artzi2013weakly,artzi2014learning,mei2015listen} represent the world indicating only the existence and direction of objects and environmental properties (Section~\ref{sec:related}). However, this representation misses the spatial relations between objects, environmental properties and the agent, such as the distance between objects, the order of objects and their relative positions with respect to the agent. To capture all the spatial information available to the agent, we propose a new grid based representation where we map the agent's view into a grid without any positional information loss (Section~\ref{representations}). We use a sequence to sequence neural network model with perceptual attention (Section~\ref{sec:architecture}) that takes advantage of the proposed perceptual representation and achieves state-of-the-art results on the SAIL dataset. % We also compare our model with strong baselines to demonstrate the abilities of different architectures (Section 5.1, 5.3). 

The SAIL dataset has certain problems regarding its size and balance.  It consists of 3237 instruction sentences with associated worlds and actions. Even though the task domain and vocabulary are fairly small, this number may be insufficient for a couple of reasons.  Typically one third of the dataset is used as a test set, and this is not large enough to differentiate models close in performance in a statistically significant manner.  Certain words or tasks occur too few times in the training set to meaningfully generalize their meaning.  In fact we often observe correct actions performed for the wrong reasons in state-of-the-art models (e.g. ``move to the chair'' always causing two steps to be taken, because that was the correct action in the only example in the training set).

Controlling the balance between different types of instructions may also be important to get a fair diagnosis of a model. Table~\ref{taskdef} shows our rough categorization of the types of tasks found in the SAIL dataset. For each type, we state the proportion of sentences in the dataset of that type, the description of the task from the perspective of the agent, and some example instructions.  One striking observation is that roughly a third of the dataset consists of ``language only'' instructions, i.e. instructions like ``turn right'' that the agent can follow without any perceptual information about the world.  In fact, our experiments show that a blind agent that does not perceive the world can guess the correct action for 64\% of the instruction sentences where the state-of-the-art performance is around 70\%.  On the other hand, semantically more complicated categories like ``move until'' may not have enough examples that will allow a learning agent to generalize correctly.  

% Solving those problems require sufficient data for each subproblem without loss of generalization and rich representations for both language and perception. The available dataset \cite{chen2011learning} is insufficient to provide enough instances to represent each problem and evaluate models on those problems.
% We detail and examine the dataset to demonstrate the inadequacies for the language grounding problem (Section 2). 

To overcome the sparse and unbalanced data problems, we propose SAILx, a synthetic data generator  (Section 3) following the recent studies on artificial data generation. Goyal et al. \shortcite{goyal2016making} and Agrawal et al. \shortcite{agrawal2016analyzing} showed that human annotated datasets may contain latent biases and neural networks are able to leverage those biases to achieve high performance. Kuhnle and Copestake \shortcite{kuhnle2017deep} and Kiela et al. \shortcite{kiela2016virtual} discussed that artificially generated data allows us to examine language understanding abilities of multimodal systems. Following this direction, our algorithmic data generation procedure provides control over the task, the language, and the world with the ability to focus on individual grounding problems.

%Conventional models developed by using this dataset lacks representing the world in terms of the spatial relations. In this study, we targeted to solve the sparse and inadequate data problem, and develop a model that benefits from the geometrical information.

The standard way of evaluating models comparing their performance on a small, fixed-sized dataset may not be adequate because of the aforementioned size and balance problems.  On the other hand, fixed-sized datasets that are too large may fail to differentiate between models of enough capacity that are all able to solve a given task.  Using a data generator we can avoid the pitfalls of fixed-sized datasets.  We can instead compare models directly in terms of their generalization power, by using the number of examples they take to learn a given task by reaching a threshold performance. We evaluate our model and compare it to others by measuring the number of instances they take to reach 90\% test set accuracy on various tasks with the SAILx generator (Section 5.2).
% The standard way of measuring improvements of a model on a task is getting better results in terms of the evaluation metric of the task. Although this approach shows the success of the model, it is not able to demonstrate the learning efficiency of the model. To test the learning efficiency of a model, we propose an evaluation metric as the number of instances that a model requires reaching a threshold which is considered as the task is solved (Section 5.2). Using this measure, we investigated the abilities of different architectures on a variety of sub-navigational instruction problems generated by our framework.
%Moreover, we can not measure the performance of models well enough on each subproblem of a task due to that a fixed-size dataset may sparsely contain instances for each sub-problem.

Our key contributions include,
\begin{itemize}
\item[-] We present a state-of-the-art model that benefits from the perceptual attention and improved word representation.
\item[-] We develop a synthetic data generation framework to overcome the small and unbalanced dataset problems.
\item[-] We present an experimental methodology to compare models in terms of learning efficiency.
\end{itemize}

\section{SAIL Dataset}\label{saildataset}

We use the SAIL dataset published by \newcite{macmahon2006walk}. In this dataset, there are 3 different maps (named \textit{Grid}, \textit{Jelly}, \textit{L}). Each map consists of different number of nodes and edges. Nodes might have an item (barstool, chair, easel, hatrack, lamp or sofa). Halls have different floor patterns (blue, brick, concrete, flower, grass, gravel, wood or yellow) and different wall paintings (butterfly, fish or tower).

The dataset was generated by using two sets of participants, six instructors and thirty-six followers. The instructors studied the mazes until they were able to efficiently navigate. Then they were asked to give a list of instructions to navigate from a starting position to a goal position without looking at the map. All written errors in the instructions, both syntactic (typos, grammar errors, etc.) and semantic (confusing left and right, calling a chair as a sofa, etc.), were kept without modification. The followers tried to follow the written list of instructions in the same maze without any prior knowledge of the environment.  They were able to complete $69.64\%$ of the paragraph length instructions accurately.

Chen and Mooney \shortcite{chen2011learning} split paragraphs into individual sentences and paired each sentence with the corresponding segment of the path. We call this version of the dataset \textit{Single-Sentence} and the original one \textit{Paragraph}. \textit{Single-Sentence} contains $3237$ instructions and \textit{Paragraph} contains $706$ sets of instructions each paired with their corresponding maps and paths.

To better understand the nature of the dataset, we analyzed the instructions in terms of their perceptual and linguistic requirements. \textit{Single-Sentence} instructions can be split into different categories: for example language only instructions such as ``turn left'' where there is no need for perception and ones such as ``turn to the chair'' which require perception. We detail the categorization and give the definition of each task in Table \ref{taskdef}. Although the dataset contains challenging language grounding problems, one-third of the data consists of language only instructions which do not require perceptual understanding, and even a model that does not use perceptual information is able to perform surprisingly well (Section 5.3). 

\begin{figure}[htpb]
    \centering
    \resizebox{0.46\textwidth}{0.2\textheight}{
    \includegraphics{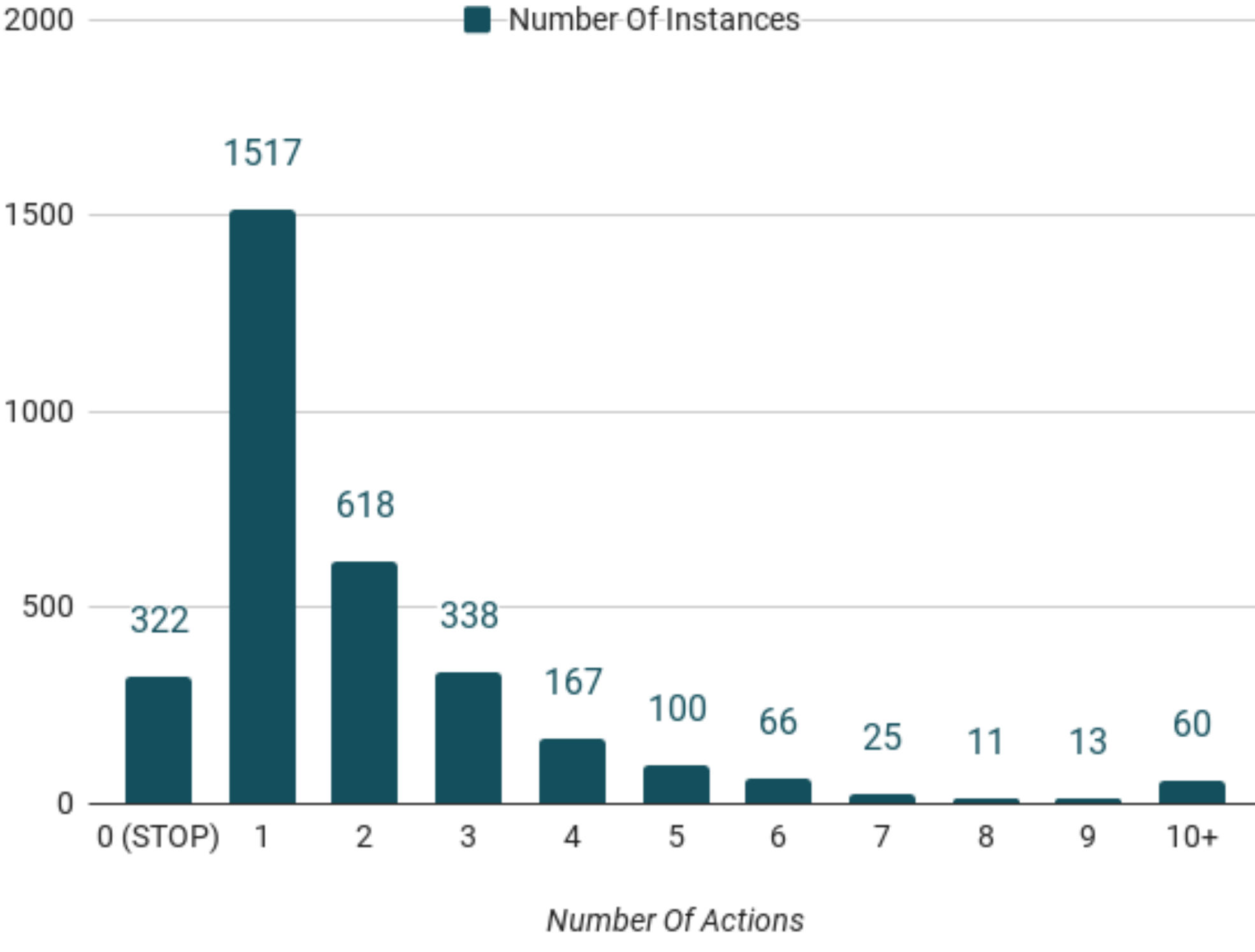}
    }
    \caption{The distribution of the length of action sequences.}
    \label{ac}
\end{figure}

Another drawback of the SAIL dataset is the lack of diversity for the action sequences that the agent should take to complete the instructions. It can be seen from Figure 2 that more than half of the instructions require a single movement followed by a stop action or only the stop action itself. % One cause of this distribution might be the poorness of maps used to collect this dataset. 

Given the problems with the dataset, machine learning models have a tendency to map specific instructions to specific action sequences due to the lack of diversity in the physical configurations (e.g. mapping "move to the chair" instruction to (MOVE, STOP) action sequence because the agent had always received that instruction when it was one step away from the chair).

\section{Synthetic Dataset (SAILx)} \label{sailx}

SAILx is a synthetic data generator we have developed that randomly generates maps, paths, and associated natural language instructions similar to the original SAIL data. 
Synthetic data generation allows us to investigate the problem of learning navigational instructions with fine-grained control over the map, the task, and the language. 
One can determine the complexity of the environment by controlling the different aspects such as the size of the map, length of the path, allowed items, flooring patterns, wall paintings and their locations. 
The language generation can be modified by controlling the vocabulary and text templates. 
% This facility allows us to test the generalization of the model in unseen configurations. 
The user chooses the category of each instance from the ones listed in Table~\ref{taskdef}. 
Thus datasets of different size and balance can be generated.
% Our data generation framework also provides a variety of isolated grounding tasks (Table 1) to focus on different challenges for the language grounding problem. 
% Most importantly, our framework provides a rich data stream combining language variety and environment complexity to train machine learning models with unlimited data.
%Although we are able to extend the language and perceptual diversity, we use the same vocabulary and visual properties with the SAIL dataset to enrich it. 
We describe the data generation procedure in this section.

\subsection{Map and Path Generation}
We generate mazes (size of 8x8, in this study) using the recursive backtracker algorithm \cite{priestley1994multipurpose} by selecting a random starting point. Once we obtain a maze, we decorate a random subset of nodes with random items. We divide the maze into two or three areas randomly and set the wall paintings of halls of each area with a distinct pattern. We choose the flooring patterns randomly but use the same pattern within a hall.  By default, we use the same set of objects, paintings and floor patterns as the SAIL dataset.  

To generate a target path we select random start and goal points that are far enough (at least four steps in this study) from each other. We find the shortest path between the two points using the $A^{*}$ algorithm \cite{hart1968formal}. If the user asks for a task pattern which does not match the generated path, we reject the path.

\subsection{Instruction Generation}
We generate instructions in the \textit{Single-Sentence} form using a large subset of the vocabulary from the original SAIL dataset. We segment a path into turning and moving parts and use one or two segments for each sentence. To generate the instructions, we use physical task patterns and pattern dependent text templates. As an example, for the \textit{Move to X} task, "reaching an end" is a physical task pattern and "/move/go/walk to the end/wall" and "take the path/hall/corridor until the end/wall" are two text templates. We select the optional parts of a template randomly. 

\subsection{Coverage}
\begin{table}[h]
\centering
\resizebox{0.48\textwidth}{!}{
\begin{tabular}{llll}
Task               & Frequency & Overall & Non-Unique \\ \hline
Language Only      & 31.7\%    & 90.25\% & 98.11\%    \\
Turn to X          & 7.01\%    & 41.41\% & 73.75\%    \\
Move to X          & 13.38\%   & 33.72\% & 60.95\%    \\
Turn and Move to X & 1.73\%    & 50.0\%  & 70.0\%     \\
Orient             & 5.16\%    & 64.67\% & 94.5\%     \\
Description        & 9.64\%    & 14.42\% & 70.83\%    \\
Move Until         & 8.71\%    & 2.84\%  & 42.86\%    \\
Any combination    & 22.67\%   & 5.31\%  & 23.86\%   
\end{tabular}
}
\caption{The task distribution and coverage statistics. The second column shows the frequency of each task in the SAIL dataset. The third column gives the percentage of the original instructions SAILx is able to generate. The last column gives the percentage for instructions that occur more than once.}
\label{freqofcats}
\end{table}

We determined the task patterns and the text templates by examining the SAIL dataset and tried to generate instructions like the ones generated by human participants. Table 2 quantifies the proportion of the original instructions that our synthetic generator is able to generate.  We are able to generate most of the instructions that occur more than once in the SAIL dataset. 
% except ones in the \textit{Any combination} category. This category contains the unique instructions and they do not share common patterns. The noise (spelling errors, typos, etc.) in the original dataset is also one of the reasons for the mismatches.

\subsection{The Fixed-Sized SAILx Dataset}
\begin{figure}[htpb]
    \centering
    \resizebox{0.4\textwidth}{0.15\textheight}{
    \includegraphics{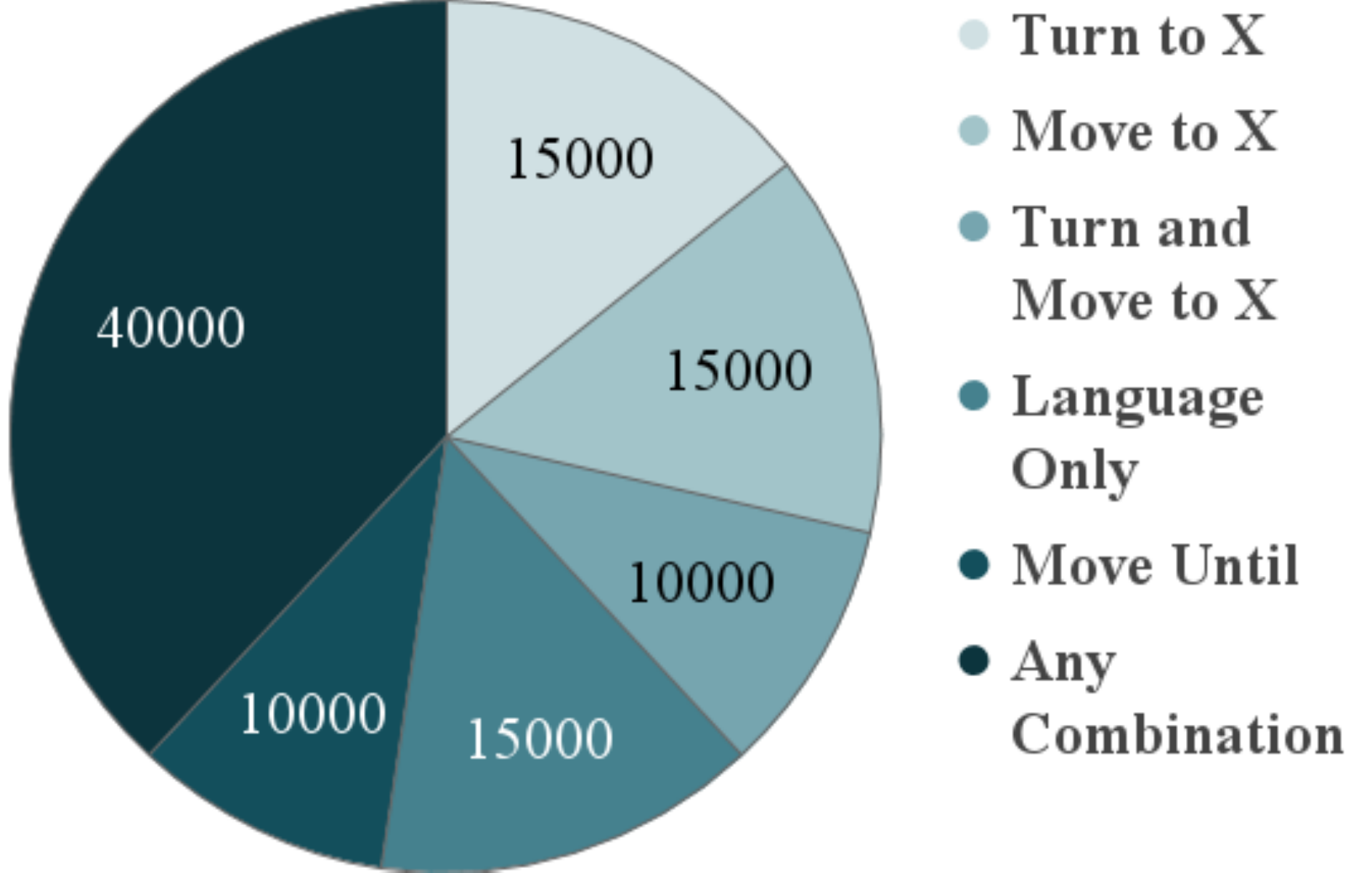}
    }
    \caption{The task distribution in the fixed-sized SAILx data.}
    \label{pie}
\end{figure}

For researchers who want to compare model performance on a fixed-sized dataset, we release a large set of pre-generated instructions with corresponding paths and maps. This dataset contains 105k instances and Figure \ref{pie} presents the proportion of each subtask. The number of instances was chosen to allow convergence to 90\% test set performance on most of the subtasks given in Table~\ref{taskdef}.  The performance of our model on this dataset is given in Section~\ref{experiments}.

\section{Model}

We use a sequence to sequence neural network model with perceptual attention to learn the meanings of navigational instructions. In this section, we first describe input representations for the natural language instructions and perceptual states. Next, we describe the architecture of the model. Finally, we outline the training and inference procedures.

\subsection{Input Representation} \label{representations}
\subsubsection*{Natural Language Instructions}
Since instructors are allowed to use free-form language, instructions may contain uppercase letters, hyphenations (e.g. \textit{blue-tiled}), shortened words (e.g. \textit{fwd} as short version of \textit{forward}) and typos (e.g. \textit{aesal} instead of \textit{easel}) in the SAIL dataset (The SAILx dataset does not contain any misspellings). Considering the small number of instructions, we split hyphenated words into subwords to prevent  sparsity. We left shortened words and typos as they are. We represented each word in the vocabulary with a one-hot vector.

\subsubsection*{Perceptual States} In previous studies, a common choice to represent the perceptual information for the agent is the concatenation of bag-of-features vectors for each direction and the agent's current position. For the SAIL dataset, the agent can observe items, flooring patterns and wall paintings of the halls. Although this representation captures the features and the directions of items, it can not capture the spatial relations.

\begin{figure}[htpb]
    \centering
    \resizebox{0.33\textwidth}{0.065\textheight}{
    \includegraphics{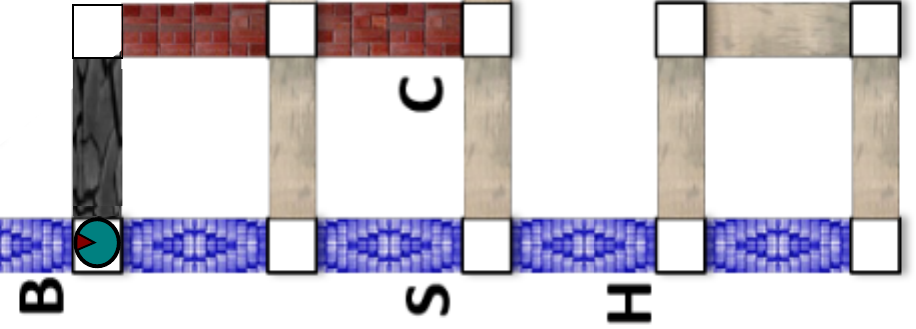}
    }
    \caption{Spatial relations between items: The circle represents the agent. The following narrative is used to describe that the agent is supposed to stay at the current position: "you should be two alleys away from a sofa and then a hatrack beyond that". To process the given instruction, the agent should understand the distance between itself and the items in the instruction, and their relative ordering.}
    \label{locality}
\end{figure}

Figure \ref{locality} illustrates the need for understanding of spatial relations. The current position of the agent is the targeted final position for a paragraph. The last instruction of that paragraph mentions this by describing the final position referring to the surrounding objects. The agent must be able to recognize its position and the relative position of the surrounding objects to understand the given instruction.

\begin{figure}[htpb]
    \centering
    \resizebox{0.38\textwidth}{0.25\textheight}{
    \includegraphics{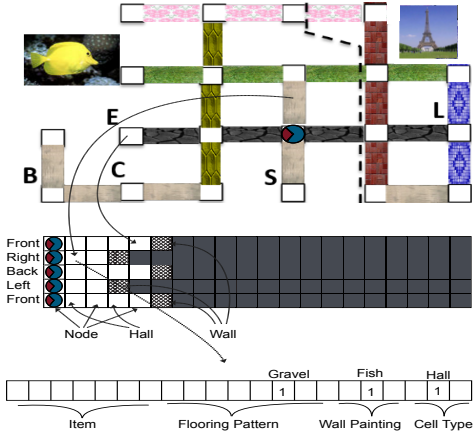}
    }
    \caption{An example grid representation for the perceptual information.}
    \label{grid}
\end{figure}

To capture the spatial properties of items and their relations with the agent, we propose a grid-based representation (Figure \ref{grid}) for perception. In this representation, the agent senses the world as a grid of cells. Each cell contains a binary vector representing either a node, a hall or a non-walkable cell. If the cell is a node, one of the item bits is set. If it is a hall the corresponding bits for the floor and wall patterns are set. In addition to material bits, one of the last three bits specifies the type of the cell. 

We always fill the grid in an agent-centric orientation. Each row contains features about a direction. And the first cell of each row always contains information about the current location of the agent. We fill the grid in a clockwise manner. We copy the first row to the last row to preserve the geometric relations. We use $20$ columns to fit the grid representation into actual maps.

\subsection{The Neural Architecture}\label{sec:architecture}
Our model\footnote{We implemented our model in julia using Knet \cite{yuret2016knet}.} is a sequence to sequence neural model \cite{sutskever2014sequence} with a perceptual attention module. It consists of three major components: an encoder, a decoder, and a convolutional neural network (CNN) with a channel attention mechanism. The encoder and the decoder model the input and output sequences by using Long Short-Term Memory (LSTM) units \cite{hochreiter1997lstm}. The CNN processes the perceptual input and the attention mechanism controls the focus on its features. Here, we give a detailed description of each component.

\begin{figure*}[htpb]
    \centering
    \resizebox{0.78\textwidth}{0.12\textheight}{
    \includegraphics{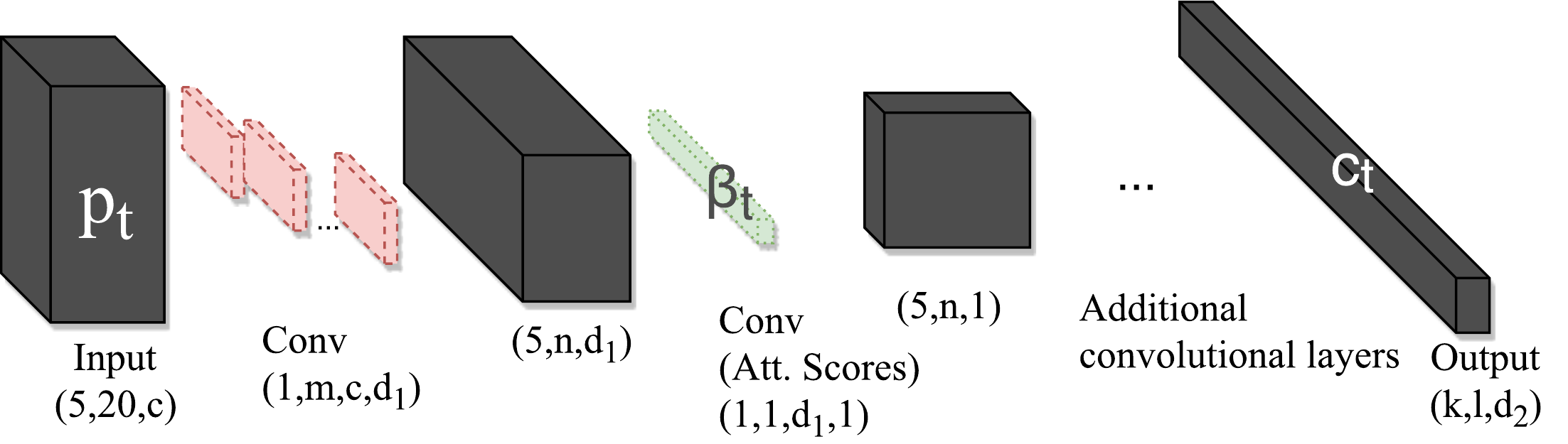}
    }
    \caption{The convolutional neural network architecture with channel attention. The perceptual input $p_t$ is a $5\times 20$ grid and $c$ is the length of the binary vector representation (Section \ref{representations}) for each grid position. The first layer applies a $(1,m,c, d_1)$ filter bank, where $1\times m$ is the size of the filter and $d_1$ is the number of filters. The output of the first layer is a $(5,n,d_1)$ tensor where $n=20-m+1$. $\beta_{t}$ is the attention vector determined by the decoder hidden state. After further convolutional layers, the output of the whole architecture, $c_t$, has $(k,l,d_2)$ dimensions.}
    \label{cnn}
\end{figure*}

\subsubsection*{Encoder}
The encoder takes the natural language instruction as a sequence of one-hot word vectors ($\textbf{w} = (w_1, w_2, ..., w_I)$) and uses a bidirectional LSTM \cite{graves2005bidirection} to process the sequence in both forward and backward directions, producing a hidden state for each position with the following formulation: 
\begin{align}%\tiny
\begin{split}
x_i & = W_e w_i\\
f_i & = \mbox{LSTM}(W_f, x_i, f_{i-1}) \\
b_i & = \mbox{LSTM}(W_b, x_i, b_{i+1}) \\
h & = f_I \oplus b_1
\end{split}
\end{align}
where $W_e$ is the embedding matrix that maps each word $w_i$ into a dense vector $x_i$, $W_f$ and $f_i$ are the parameters and the hidden state for the forward LSTM, $W_b, b_i$ are the parameters and the hidden state for the backward LSTM, $h$ is the final hidden state for the encoder and is obtained by concatenating  $f_I$ and $b_1$. We use zero vectors for the initial $f_0$ and $b_{I+1}$.

\subsubsection*{A CNN with channel attention}
We designed the CNN architecture (see Figure \ref{cnn}) such that the first layer of the network would be able to detect objects/properties on the perceptual input $p_t$ by applying a filter bank. Then, a decoder-controlled attention vector weighs the properties captured on the output channels of this first layer.  The purpose of the attention vector is to direct perception to parts of the input relevant to the instruction.
%by using the predicted scores as a filter. 
% The decoder of a regular \cite{sutskever2014sequence} sequence to sequence model tries to produce output sequence from the encoded input sequence by only using the final hidden state of the encoder and the input to the decoder at each time step. This architecture suffers from the lack of a direct link between language and the perceptual information. To create this connection, we propose an attention mechanism that learns a filter to select the important features from the first layer of the CNN by attending the channels. 
We formulate the attention vector as follows:
\begin{align}
\begin{split}
\beta_{ti} & = \frac{\exp(e_{ti})}{\sum_{j=1}^{I}\exp(e_{tj})}
\end{split}
\end{align}
where $\beta_{ti}$ is the weight of input channel $i$ at time $t$.  $\beta_{ti}$ is calculated as the softmax of an attention score $e_{ti}$ which is a function of the hidden state of the decoder $s_{t-1}$:
\begin{align}
e_{ti} = \mbox{NN}(W_a, s_{t-1})
\end{align}
We use a feedforward neural network NN with parameters $W_a$ and $s_{t-1}$ as the input for the computation of the attention score.

After the attention layer, we have further convolutional layers\footnote{We use one additional layer for SAILx dataset and two additional layers for SAIL dataset} to learn higher order relations or make comparisons among different directions. We apply a relu activation after each convolution layer except the last one, where we use the sigmoid function, giving the final perceptual state $c_t$.

\subsubsection*{Decoder}
The decoder generates a sequence of actions $\boldsymbol{a} = (a_{1}, a_{2}, .., a_{T})$ given the perceptual state $c_{t}$ and the language state $h$. The decoder defines a probability over the generated action sequence $\boldsymbol{a}$ as follows:
\begin{align}%\tiny
P(\boldsymbol{a}) = \prod_{t=1}^{T}P(a_{t} | s_t, c_t)
\end{align}
Our architecture models each conditional probability using the following formulation:

\begin{align}
\begin{split}\tiny
& c_{t} = \mbox{CNN}(W_c, p_t, \beta_{t}) \\
& s_{t} = \mbox{LSTM}(W_d, c_t \oplus a_{t-1}, s_{t-1})\\
& o_{t} = W_{1}s_{t} + W_{2}c_{t} + b\\
& P(a_{t} | s_t, c_t) = \mbox{softmax}(o_{t})
\end{split}
\end{align}
where $p_t$ is the perceptual state at time step $t$, $c_{t}$ is the output of the CNN with parameters $W_c$ and the attention distribution $\beta_{t}$. $s_{t}$ is the hidden state of the decoder LSTM with parameters $W_d$. The input of the LSTM is the concatenation of $c_t$ and $a_{t-1}$.  The unnormalized output $o_t$ is obtained by a linear combination of $c_t$ and $s_t$.  $o_t$ is normalized by the softmax operation to determine the conditional probabilities for possible actions.

\subsection{Training}
We use the cross-entropy loss to maximize the probability of the ground-truth action sequence. If the correct action sequence is $[a_{1}, a_{2}, ..., a_{T}]$, we can train the model by minimizing the negative log-likelihood:
\begin{align}%\tiny
L &= -\log P(a_{1}, a_{2}, ..., a_{T} | \textbf{w},p_{1:T})\\
  &= -\log\prod_{t=1}^{T}P(a_{t} | s_t, c_t)\\
  &= -\sum_{t=1}^{T}\log P(a_{t} | s_t, c_t)
\end{align}
Since we model conditional probabilities with a differentiable recurrent model (3), weights of the model can be learned with backpropagation through time \cite{werbos1990backpropagation}.

\subsection{Inference}
Once we train the model, we generate action sequences by searching over alternative paths with beam search \cite{sutskever2014sequence,rush2015neural,mei2015listen} using the distribution $P(a_{t} | s_t, c_t)$. For the \textit{Paragraph} instances, we execute the beam search sentence by sentence, keeping the same beam between sentences. 
%When the search is finished for one sentence, we start the next sentence with the same beam. 
%We start the search for each sentence in the paragraph by feeding the instruction to the encoder and setting the hidden state of the decoder as the final state of the encoder. 
We generate actions for a sentence until the \textit{STOP} action is taken, the agent hits a wall, or reaches a maximum number of actions. 
%We feed the representation of the final position of each sequence as the first input of the decoder for each branch. 
We also use an ensemble of trained models to obtain the action distribution by taking the average of the $P(a_{t} | s_t, c_t)$ distributions predicted by each model.

\section{Experiments \& Results}\label{experiments}
We conducted experiments on both the SAIL and the SAILx datasets. First, we describe the baseline models used in the experiments. Then, we give the details of the experiments on the SAIL dataset and compare our model with the state-of-the-art. Finally, we explain our experiments on the SAILx dataset and the propose evaluation metric for model comparison.

\subsection{Baselines}

\textbf{Language Only (L.O.).} This model is a regular encoder-decoder architecture without an attention mechanism. The encoder encodes the instruction with a bidirectional LSTM and the decoder predicts the action sequence taking the previous action as input at each time step. We feed the \textit{STOP} action to the decoder as the initial input. This model does not use any perceptual information. We use gold actions as the input of the decoder during training. At test time, we use the actions predicted by the decoder.

\noindent\textbf{Bag-Of-Features (B.O.F).} This baseline model is also a regular encoder-decoder model with a bag-of-features representation (Section 4.1) for world states. The encoder encodes the instruction with a bidirectional LSTM and the decoder predicts the action sequence taking the bag-of-features world state as input at each time step.

\subsection{SAIL dataset}
We experimented on both the \textit{Single-Sentence} and the \textit{Paragraph} version of the SAIL dataset. Following the previous studies, a trial is counted as successful if and only if the final position and orientation match with the ground-truth path for the \textit{Single-Sentence} version. For the \textit{Paragraph} version, matching the final position is sufficient for success.

We use the \textit{Single-Sentence} version of the data for the training and we test the model on both \textit{Single-Sentence} and \textit{Paragraph} datasets. We use six fold cross validation for the tuning of the model using splits for each map (\textit{Grid, Jelly, L}). We left two maps as training data and split the remaining one into development data (50\%) and test data (50\%). We repeat this experiment by swapping the test and development data. This process is carried out for each map as development/test data. We run each fold ten times and report the size-weighted average of runs as the final score. We tune the hyper-parameters of the model depending on the average score of the six folds instead of tuning for each map.

The common approach on this dataset is using the test data for development, and denoted by ``vTest''. This approach is problematic because of the usage of the test data in the tuning process. Mei et al. \shortcite{mei2015listen} proposed another approach (vDev) to tune the model by using 10\% of the two maps as the development data and the remaining (90\%) as the training data. This approach is also problematic because there may be linguistic differences among the instructors for different maps. By using half of the test map data for development and the other half for test, we avoid both problems.

We use Adam \cite{kingma2014adam} with default parameters for the optimization and gradient clipping \cite{mikolov2010} with the norm threshold $5$. We use the success rate of the development data for early stopping. We stop the training if the model does not improve the development score within $10$ iterations. We update the model after seeing each paired instruction and action sequence.

\begin{table}[ht]
\centering
\resizebox{\columnwidth}{!}{
\begin{tabular}{lll}
\hline
Method                        & Single-Sentence & Paragraph \\ \hline
Chen and Mooney \shortcite{chen2011learning}       & 54.40           & 16.18     \\
Chen \shortcite{chen2012fast}                   & 57.28           & 19.18     \\
Kim and Mooney \shortcite{kim2012unsupervised}         & 57.22           & 20.17     \\
Kim and Mooney \shortcite{kim2013adapting}         & 62.81           & 26.57     \\
Artzi and Zettlemoyer \shortcite{artzi2013weakly}  & 65.28 & 31.93 \\
Artzi et al. \shortcite{artzi2014learning} & 64.36 & \textbf{35.44}     \\
Andreas and Klein \shortcite{andreas2015alignment}      & 59.60           & -         \\
Ko{\v{c}}isk{\`y} et al. \shortcite{kovcisky2016semantic} (vDev)             & 63.25           & -     \\
Mei et al. \shortcite{mei2015listen} (vDev) (ens=10)             & 69.98           & 26.07     \\
Mei et al. \shortcite{mei2015listen} (ens=10)            & 71.05           & 30.34     \\
Human \cite{macmahon2006walk} & - & 69.64     \\
\hline
Our Model (avg) & 68.53 & 25.51 \\
Our Model (ens=10) & \textbf{72.82} & 32.57\\
\hline
\end{tabular}
}
\caption{Overall results. Results of the previous works were obtained using the ``vTest'' procedure if not explicitly stated otherwise.}
\label{mainresults}
\end{table}

Table \ref{mainresults} shows the overall performance of our model against the previous studies with the ``vTest'' experimenting scheme. Our final model achieved the state-of-the-art result on \textit{Single-Sentence} even though we do not tune the model for each map and use the test set for tuning. The performance on the \textit{Paragraph} dataset is comparable with the best model \cite{artzi2014learning} which uses an additional semantic lexicon for the initialization of a semantic parser.
%Although previous work uses an attention onto words, we observed that adding this mechanism does not bring any improvements. We suspect that initializing decoder with encoder's last hidden states enough to convey the lingual information.

\begin{table}[h]
\centering
\begin{tabular}{llll}
Task               & L.O.    & B.O.F.  & Ours \\ \hline
Language Only      & 86.65 & 87.23 & 89.38    \\
Turn to X          & 85.02 & 87.66 & 88.11    \\
Move to X          & 62.82 & 73.21 & 76.91    \\
Turn and Move to X & 44.64 & 64.29 & 73.21    \\
Orient             & 92.22 & 93.41 & 92.81    \\
Description           & 86.54 & 86.22 & 89.42    \\
Move Until         & 38.3  & 47.87 & 46.45    \\
Any combination    & 20.16 & 33.24 & 35.56    \\ \hline
Overall            & 63.61 & 69.54 & 71.58   
\end{tabular}
\caption{The performance (test accuracy) comparison of baselines and the proposed model. Results were obtained using ensemble of three models for each architecture.}
\label{comparison}
\end{table}

Table \ref{comparison} presents detailed results of baseline models compared to our final model. The \emph{Language Only} baseline achieves surprisingly good results on the visual tasks without receiving perceptual information. One reason for this is that some instructions for the visual tasks contain non-perceptual clues to solve the task, e.g. \textit{"turn \textbf{right} to the chair", "move \textbf{forward two steps} to the easel"}. The ability of the network to exploit the bias hidden in the action sequence distribution (see Figure \ref{ac}) might be another factor for the high performance of the language only model. For example, most of the action sequences are (\textit{RIGHT, STOP}) in the \textit{Orient} task. 

The Bag-Of-Features baseline performs better than the Language Only baseline in almost all tasks. Since both models have the same architecture for instruction modeling, it is expected that they result in similar performance in the Language Only task which we observe. Our final model further improves the B.O.F baseline in most tasks.

\subsection{SAILx dataset}
We use our synthetic dataset to compare the learning efficiencies of the proposed grid based model and the bag-of-features baseline. We train each model on a stream of data until the test accuracy exceeds a threshold. We use Adam \cite{kingma2014adam} with default parameters for the optimization. We use gradient clipping \cite{mikolov2010} and set the norm of the gradients $5$.

\begin{figure}[htpb]
    \centering
    \resizebox{0.5\textwidth}{0.35\textheight}{
    \includegraphics{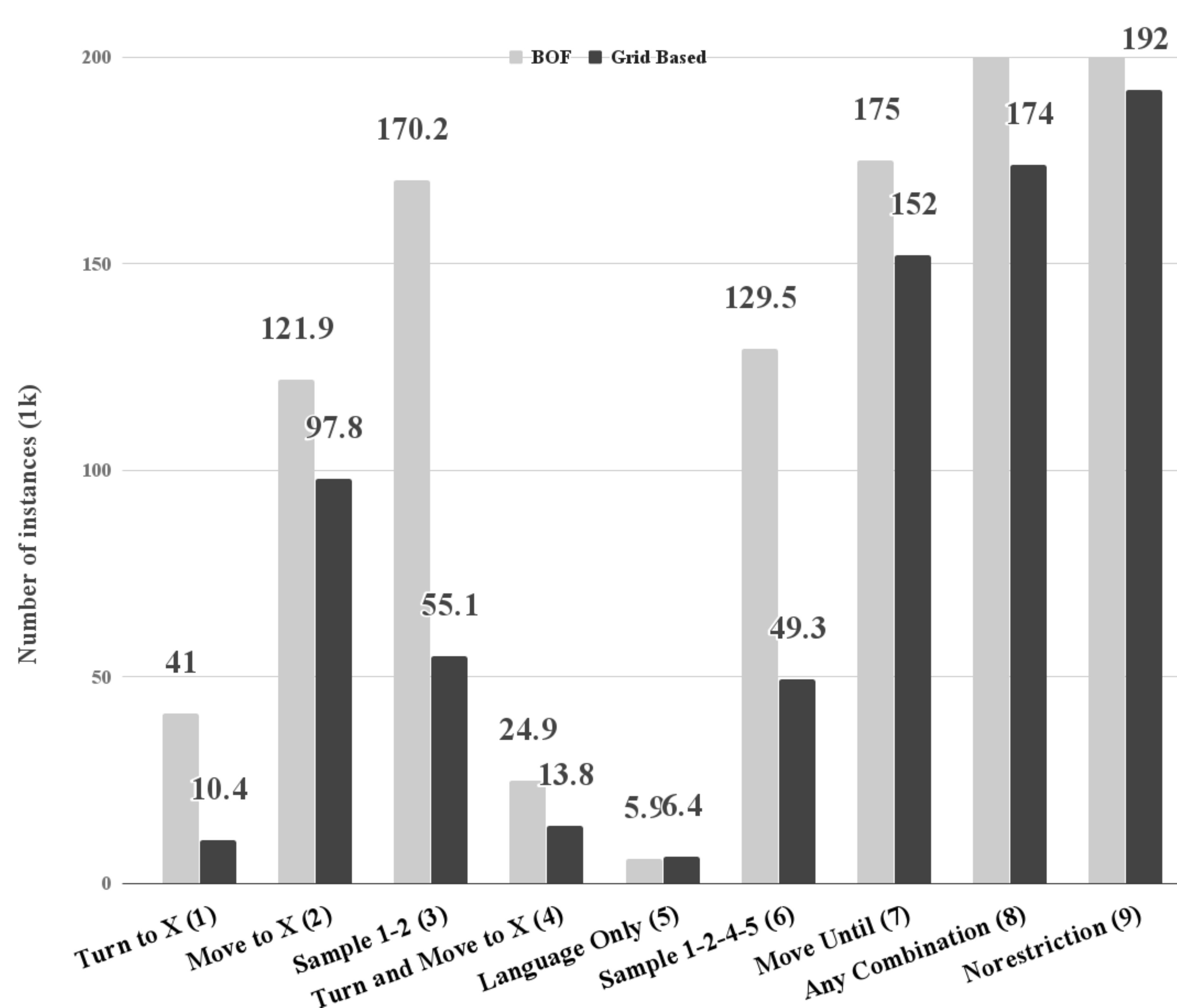}
    }
    \caption{The learning efficiency comparison of the B.O.F baseline and the grid based model. The y axis shows the number of instances required to reach 90\% accuracy on unseen data.}
    \label{synthetic}
\end{figure}

Figure \ref{synthetic} demonstrates the number of instances required to reach 90\% test accuracy by the B.O.F. baseline and our final model\footnote{The test accuracy is calculated as a moving average of accuracy on the next unseen instance: $moving\_average = 0.95*moving\_average + 0.05*accuracy\_on\_current\_batch$.}. We did not include the L.O. baseline in this figure because it does not reach 90\% on most of the tasks.  
%Since the B.O.F model does not capture spatial relations, it is not able to solve the tasks that involve spatial relations without random exploration. 
%Hence, we selected the threshold by observing the minimum of maximum test accuracy that the B.O.F model reached in all task categories. 
We performed the hyper-parameter search for each model and for each task using golden section search \cite{kiefer1953sequential}.

Our grid based model achieves better or equal efficiency in almost all tasks when it is compared to the performance of the B.O.F baseline. The B.O.F baseline is not able to solve Any Combination and Norestriction tasks within the 250k instances limit. Additionally, both models show similar performance on the \textit{Language Only} task because of their similar architecture for language processing.

\begin{table}[ht]
\centering
\resizebox{0.8\columnwidth}{!}{
\begin{tabular}{lllll}
      & \multicolumn{2}{c}{Dev}                    & \multicolumn{2}{c}{Test} \\
      & Avg. & \multicolumn{1}{l|}{Ens.} & Avg.   & Ens.  \\ \cline{2-5} 
L.O   & 54.85     & \multicolumn{1}{l|}{54..41}    & 55.03       & 54.64      \\
B.O.F & 84.9      & \multicolumn{1}{l|}{86.04}     & 84.61       & 85.42      \\
Ours  & 91.93     & \multicolumn{1}{l|}{93.69}     & 91.82       & 93.48     
\end{tabular}
}
\caption{Results on the generated fixed data. The average and ensemble scores were obtained by using 10 models for each architecture.}
\label{sailxexps}
\end{table}

In Table \ref{sailxexps} we compare the performance of the baseline models and the grid based model on the fixed-sized SAILx dataset (Section \ref{sailx}). We split this data into train (70\%), dev (15\%) and test (15\%) sets while preserving the proportion of each subtask. The grid based model is significantly better than the B.O.F baseline. The L.O baseline reaches an accuracy higher than the proportion of the language only data. One reason might be that since the model uses the previous action as the input of the decoder, it captures some action sequence patterns.

\section{Related Work}\label{sec:related}
In this section, we first summarize the studies on artificial data generation. Next, we outline the general literature on grounded language learning.

\subsection*{Artificial Data Generation}
% Data eager nature of neural networks requires a tremendous amount of data for better generalization. Artificial data generation is an easy way of providing this amount of data for detailed examination. Bengio et al. \shortcite{bengio1994learning}, Hochreiter and Schmidhuber \shortcite{hochreiter1997lstm}, Joulind and Mikolov \shortcite{joulin2015inferring} and Kaiser and Sutskever \shortcite{kaiser2015neural} have examined their systems on the algorithmic tasks. Similarly, combinatorial problems \cite{vinyals2015pointer}, atari games \cite{mnih2013playing}, code execution \cite{zaremba2014learning} have been used as a test bed for neural networks.

%Developing multi-modal datasets for machine learning might be problematic due to hidden biases in the language or the requirement of a complex world for complex language generation. Goyal et al. \shortcite{goyal2016making} and Agraval et al. \shortcite{agrawal2016analyzing} have shown that deep neural networks are able to exploit the hidden biases of human annotated visual question answering (VQA) datasets. 
Similar to SAILx, SHAPEWORLD \cite{kuhnle2017shapeworld}, SHAPES \cite{andreas2016neural}, CLEVR \cite{johnson2016clevr} datasets provide control over the language and the world to examine subproblems of the visual question answering problem. bAbI tasks \cite{weston2015towards} also provide controlled subtasks, however, it is text only whereas our synthetic dataset is multi-modal.

Yu et al. \shortcite{yu2017deep} propose an environment for navigating in a 2-D maze like environment, XWORLD. In contrast to the SAIL environment, in XWORLD, the environment is fully observable to the agent. \shortcite{chaplot2017gated} propose a new 3D environment built on the ViZDoom API \cite{kempka2016vizdoom}. They manually generated 70 navigation instructions where each instruction is a combination of (action, attribute(s), object) triple. \shortcite{hermann2017grounded}  propose a simulated environment in 3-D. 
%In this environment, the agent is instructed to go to a specific object which is described using referring expressions. 
The language used in both environments involve only referring expressions to navigate the agent. 
%In our framework, the language is generated from free-form text templates, which are structured sentences and include action verbs.

\subsection*{Grounded Language Learning}
% Winograd's SHRDLU was an early work in grounded language understanding which integrated perceptual information with language manually in a blocks world scenario \cite{winograd1973procedural}. Nilsson \shortcite{nilsson1984shakey} connected the perception with the symbolic representation in the robot Shakey, for the first time. More recently, \cite{siskind1993naive,siskind1995grounding,siskind2001grounding} built temporal relations between objects observed visually and provided a basis for the semantic roles of the participants for verbs. Roy \shortcite{roy2005grounding} showed that the symbolic modeling is inadequate to capture the context dependent word meaning and proposed to use the first-person perspective sensory data in grounded learning.

MacMahon et al. \shortcite{macmahon2006walk} published the SAIL dataset and demonstrated a rule based system that uses manually aligned language and world features. The most studied approach in this domain is mapping natural language instructions into a formal meaning representation. Chen and Mooney \shortcite{chen2011learning} presented a system that translates instructions to formal executable plans by training the KRISP semantic parser \cite{kate2006using} with aligned instruction and action-sequence pairs. Later, Chen \shortcite{chen2012fast} improved the system by modifying the underlying semantic grammar and tested the system also on Chinese instructions. Kim and Mooney \shortcite{kim2012unsupervised} approached grounded language learning as probabilistic context free grammar (PCFG) induction introduced by Borschinger \shortcite{borschinger2011reducing} to the navigation domain. They further improve this technique using a re-ranking module with the task-specific weak signal \cite{kim2013adapting}. Artzi and Zettlemoyer \shortcite{artzi2013weakly} learned a semantic parser seeded with a manual lexicon. The parser operates on a combinatory categorical grammar (CCG) to translate the instruction into a lambda-calculus formalism for the semantic representation. They improved their system using a re-ranker and controlling the size of the lexicon in a data driven fashion \cite{artzi2014learning}.

Another approach is learning to translate the instruction into an action sequence in an end-to-end fashion. Andreas and Klein \shortcite{andreas2015alignment} modeled the instruction following task as scoring possible execution plans depending on the alignment with a given instruction. They use a conditional random field model to learn this alignment. Mei et al. \shortcite{mei2015listen} use a textual attention-based \cite{bahdanau2014neural} encoder decoder neural network in contrast to our model which has an attention mechanism over the channels for the visual attributes. Although previous work uses attention to words, we observed that adding this mechanism does not bring any improvements. We suspect that initializing the decoder with the encoder's last hidden state is enough to convey the linguistic information in the SAIL domain. Our grid based representation for the perceptual states brings an improvement over the bag-of-features representation of \cite{mei2015listen}. Ko{\v{c}}isk{\`y} et al. \shortcite{kovcisky2016semantic} utilizes an auto-encoder objective to enable semi-supervised training by leveraging randomly generated unsupervised data (random action sequences). They use a similar model to the bag-of-features baseline and show that the model is able to benefit from unsupervised training.

\section{Conclusion}
We have developed a synthetic data generation framework that can be used to provide an unlimited dataset perception-instruction-action instances to train and evaluate grounded language learning models. We proposed an evaluation metric to measure the learning efficiency of a model using the number of instances to reach a particular performance rather than the accuracy reached on a fixed-sized dataset.  We developed a novel grid based representation for the perceptual states where spatial relations are missing in previous approaches. Our model resulted in state-of-the-art accuracy in \textit{Single-Sentence} and achieved comparable results in \textit{Paragraph} without using any external resources.

% Even though our model produces an action sequence for a given instruction, it does not incorporate the whole action sequence in the training. Global normalization \cite{weiss2015structured,wiseman2016sequence,lee2016global} method models the probability of the whole sequence instead of the combination of the local decisions with the help of beam search, which may bring improvements thanks to the exploration of the world during the training. Additionally, incorporating the beam search into the training allows the \textit{Paragraph} level training which may improve the performance of the model on the consecutive instructions that contain referring expressions (e.g. turn to the chair. move to it.).

% In our navigation domain, perceptual input consists of discrete features which provide a lossless representation. However, working on raw pixels allows more complex inputs and challenging problems. For example, the meaning of the preposition \textit{"on"} changes depending on the context, \emph{"move forward to the lamp on the wall"} brings a horizontal relation whereas \emph{"move forward to the lamp on the table"} indicates the horizontal relation. We are planning to extend our synthetic data generation methodology to the 3-D world and work on the more challenging language components. 

\section*{Acknowledgements}
This work was supported by the Scientific and Technological Research Council of Turkey (TUBITAK) grants 114E628 and 215E201.

\bibliography{acl2017}

\begin{thebibliography}{}

\bibitem[\protect\citename{Agrawal \bgroup et al.\egroup
  }2016]{agrawal2016analyzing}
Aishwarya Agrawal, Dhruv Batra, and Devi Parikh.
\newblock 2016.
\newblock Analyzing the behavior of visual question answering models.
\newblock {\em arXiv preprint arXiv:1606.07356}.

\bibitem[\protect\citename{Andreas and Klein}2015]{andreas2015alignment}
Jacob Andreas and Dan Klein.
\newblock 2015.
\newblock Alignment-based compositional semantics for instruction following.
\newblock {\em arXiv preprint arXiv:1508.06491}.

\bibitem[\protect\citename{Andreas \bgroup et al.\egroup
  }2016]{andreas2016neural}
Jacob Andreas, Marcus Rohrbach, Trevor Darrell, and Dan Klein.
\newblock 2016.
\newblock Neural module networks.
\newblock In {\em Proceedings of the IEEE Conference on Computer Vision and
  Pattern Recognition}, pages 39--48.

\bibitem[\protect\citename{Artzi and Zettlemoyer}2013]{artzi2013weakly}
Yoav Artzi and Luke Zettlemoyer.
\newblock 2013.
\newblock Weakly supervised learning of semantic parsers for mapping
  instructions to actions.
\newblock {\em Transactions of the Association for Computational Linguistics},
  1:49--62.

\bibitem[\protect\citename{Artzi \bgroup et al.\egroup
  }2014]{artzi2014learning}
Yoav Artzi, Dipanjan Das, and Slav Petrov.
\newblock 2014.
\newblock Learning compact lexicons for ccg semantic parsing.
\newblock In {\em EMNLP}, pages 1273--1283.

\bibitem[\protect\citename{Bahdanau \bgroup et al.\egroup
  }2014]{bahdanau2014neural}
Dzmitry Bahdanau, Kyunghyun Cho, and Yoshua Bengio.
\newblock 2014.
\newblock Neural machine translation by jointly learning to align and
  translate.
\newblock {\em arXiv preprint arXiv:1409.0473}.

\bibitem[\protect\citename{B{\"o}rschinger \bgroup et al.\egroup
  }2011]{borschinger2011reducing}
Benjamin B{\"o}rschinger, Bevan~K Jones, and Mark Johnson.
\newblock 2011.
\newblock Reducing grounded learning tasks to grammatical inference.
\newblock In {\em Proceedings of the Conference on Empirical Methods in Natural
  Language Processing}, pages 1416--1425. Association for Computational
  Linguistics.

\bibitem[\protect\citename{Chaplot \bgroup et al.\egroup
  }2017]{chaplot2017gated}
Devendra~Singh Chaplot, Kanthashree~Mysore Sathyendra, Rama~Kumar Pasumarthi,
  Dheeraj Rajagopal, and Ruslan Salakhutdinov.
\newblock 2017.
\newblock Gated-attention architectures for task-oriented language grounding.
\newblock {\em arXiv preprint arXiv:1706.07230}.

\bibitem[\protect\citename{Chen and Mooney}2011]{chen2011learning}
David~L Chen and Raymond~J Mooney.
\newblock 2011.
\newblock Learning to interpret natural language navigation instructions from
  observations.
\newblock In {\em AAAI}, volume~2, pages 1--2.

\bibitem[\protect\citename{Chen}2012]{chen2012fast}
David~L Chen.
\newblock 2012.
\newblock Fast online lexicon learning for grounded language acquisition.
\newblock In {\em Proceedings of the 50th Annual Meeting of the Association for
  Computational Linguistics: Long Papers-Volume 1}, pages 430--439. Association
  for Computational Linguistics.

\bibitem[\protect\citename{Goyal \bgroup et al.\egroup }2016]{goyal2016making}
Yash Goyal, Tejas Khot, Douglas Summers-Stay, Dhruv Batra, and Devi Parikh.
\newblock 2016.
\newblock Making the v in vqa matter: Elevating the role of image understanding
  in visual question answering.
\newblock {\em arXiv preprint arXiv:1612.00837}.

\bibitem[\protect\citename{Graves and Schmidhuber}2005]{graves2005bidirection}
Alex Graves and J{\"u}rgen Schmidhuber.
\newblock 2005.
\newblock Framewise phoneme classification with bidirectional lstm and other
  neural network architectures.
\newblock {\em Neural Networks}, 18(5):602--610.

\bibitem[\protect\citename{Hart \bgroup et al.\egroup }1968]{hart1968formal}
Peter~E Hart, Nils~J Nilsson, and Bertram Raphael.
\newblock 1968.
\newblock A formal basis for the heuristic determination of minimum cost paths.
\newblock {\em IEEE transactions on Systems Science and Cybernetics},
  4(2):100--107.

\bibitem[\protect\citename{Hermann \bgroup et al.\egroup
  }2017]{hermann2017grounded}
Karl~Moritz Hermann, Felix Hill, Simon Green, Fumin Wang, Ryan Faulkner, Hubert
  Soyer, David Szepesvari, Wojtek Czarnecki, Max Jaderberg, Denis Teplyashin,
  et~al.
\newblock 2017.
\newblock Grounded language learning in a simulated 3d world.
\newblock {\em arXiv preprint arXiv:1706.06551}.

\bibitem[\protect\citename{Hochreiter and Schmidhuber}1997]{hochreiter1997lstm}
Sepp Hochreiter and J{\"u}rgen Schmidhuber.
\newblock 1997.
\newblock Long short-term memory.
\newblock {\em Neural computation}, 9(8):1735--1780.

\bibitem[\protect\citename{Johnson \bgroup et al.\egroup
  }2016]{johnson2016clevr}
Justin Johnson, Bharath Hariharan, Laurens van~der Maaten, Li~Fei-Fei,
  C~Lawrence Zitnick, and Ross Girshick.
\newblock 2016.
\newblock Clevr: A diagnostic dataset for compositional language and elementary
  visual reasoning.
\newblock {\em arXiv preprint arXiv:1612.06890}.

\bibitem[\protect\citename{Kate and Mooney}2006]{kate2006using}
Rohit~J Kate and Raymond~J Mooney.
\newblock 2006.
\newblock Using string-kernels for learning semantic parsers.
\newblock In {\em Proceedings of the 21st International Conference on
  Computational Linguistics and the 44th annual meeting of the Association for
  Computational Linguistics}, pages 913--920. Association for Computational
  Linguistics.

\bibitem[\protect\citename{Kempka \bgroup et al.\egroup
  }2016]{kempka2016vizdoom}
Micha{\l} Kempka, Marek Wydmuch, Grzegorz Runc, Jakub Toczek, and Wojciech
  Ja{\'s}kowski.
\newblock 2016.
\newblock Vizdoom: A doom-based ai research platform for visual reinforcement
  learning.
\newblock In {\em Computational Intelligence and Games (CIG), 2016 IEEE
  Conference on}, pages 1--8. IEEE.

\bibitem[\protect\citename{Kiefer}1953]{kiefer1953sequential}
Jack Kiefer.
\newblock 1953.
\newblock Sequential minimax search for a maximum.
\newblock {\em Proceedings of the American mathematical society},
  4(3):502--506.

\bibitem[\protect\citename{Kiela \bgroup et al.\egroup }2016]{kiela2016virtual}
Douwe Kiela, Luana Bulat, Anita~L Vero, and Stephen Clark.
\newblock 2016.
\newblock Virtual embodiment: A scalable long-term strategy for artificial
  intelligence research.
\newblock {\em arXiv preprint arXiv:1610.07432}.

\bibitem[\protect\citename{Kim and Mooney}2012]{kim2012unsupervised}
Joohyun Kim and Raymond~J Mooney.
\newblock 2012.
\newblock Unsupervised pcfg induction for grounded language learning with
  highly ambiguous supervision.
\newblock In {\em Proceedings of the 2012 Joint Conference on Empirical Methods
  in Natural Language Processing and Computational Natural Language Learning},
  pages 433--444. Association for Computational Linguistics.

\bibitem[\protect\citename{Kim and Mooney}2013]{kim2013adapting}
Joohyun Kim and Raymond~J Mooney.
\newblock 2013.
\newblock Adapting discriminative reranking to grounded language learning.
\newblock In {\em ACL (1)}, pages 218--227.

\bibitem[\protect\citename{Kingma and Ba}2014]{kingma2014adam}
Diederik Kingma and Jimmy Ba.
\newblock 2014.
\newblock Adam: A method for stochastic optimization.
\newblock {\em arXiv preprint arXiv:1412.6980}.

\bibitem[\protect\citename{Ko{\v{c}}isk{\`y} \bgroup et al.\egroup
  }2016]{kovcisky2016semantic}
Tom{\'a}{\v{s}} Ko{\v{c}}isk{\`y}, G{\'a}bor Melis, Edward Grefenstette, Chris
  Dyer, Wang Ling, Phil Blunsom, and Karl~Moritz Hermann.
\newblock 2016.
\newblock Semantic parsing with semi-supervised sequential autoencoders.
\newblock {\em arXiv preprint arXiv:1609.09315}.

\bibitem[\protect\citename{Kuhnle and Copestake}2017a]{kuhnle2017deep}
Alexander Kuhnle and Ann Copestake.
\newblock 2017a.
\newblock Deep learning evaluation using deep linguistic processing.
\newblock {\em arXiv preprint arXiv:1706.01322}.

\bibitem[\protect\citename{Kuhnle and Copestake}2017b]{kuhnle2017shapeworld}
Alexander Kuhnle and Ann Copestake.
\newblock 2017b.
\newblock Shapeworld-a new test methodology for multimodal language
  understanding.
\newblock {\em arXiv preprint arXiv:1704.04517}.

\bibitem[\protect\citename{MacMahon \bgroup et al.\egroup
  }2006]{macmahon2006walk}
Matt MacMahon, Brian Stankiewicz, and Benjamin Kuipers.
\newblock 2006.
\newblock Walk the talk: Connecting language, knowledge, and action in route
  instructions.
\newblock {\em Def}, 2(6):4.

\bibitem[\protect\citename{Mei \bgroup et al.\egroup }2015]{mei2015listen}
Hongyuan Mei, Mohit Bansal, and Matthew~R Walter.
\newblock 2015.
\newblock Listen, attend, and walk: Neural mapping of navigational instructions
  to action sequences.
\newblock {\em arXiv preprint arXiv:1506.04089}.

\bibitem[\protect\citename{Mikolov \bgroup et al.\egroup }2010]{mikolov2010}
Tomas Mikolov, Martin Karafi{\'a}t, Lukas Burget, Jan Cernock{\`y}, and Sanjeev
  Khudanpur.
\newblock 2010.
\newblock Recurrent neural network based language model.
\newblock In {\em INTERSPEECH 2010, 11th Annual Conference of the International
  Speech Communication Association, Makuhari, Chiba, Japan, September 26-30,
  2010}, pages 1045--1048.

\bibitem[\protect\citename{Priestley and Ward}1994]{priestley1994multipurpose}
Hilary~A Priestley and Martin~P Ward.
\newblock 1994.
\newblock A multipurpose backtracking algorithm.
\newblock {\em Journal of Symbolic Computation}, 18(1):1--40.

\bibitem[\protect\citename{Rush \bgroup et al.\egroup }2015]{rush2015neural}
Alexander~M Rush, Sumit Chopra, and Jason Weston.
\newblock 2015.
\newblock A neural attention model for abstractive sentence summarization.
\newblock {\em arXiv preprint arXiv:1509.00685}.

\bibitem[\protect\citename{Sutskever \bgroup et al.\egroup
  }2014]{sutskever2014sequence}
Ilya Sutskever, Oriol Vinyals, and Quoc~V Le.
\newblock 2014.
\newblock Sequence to sequence learning with neural networks.
\newblock In {\em Advances in neural information processing systems}, pages
  3104--3112.

\bibitem[\protect\citename{Werbos}1990]{werbos1990backpropagation}
Paul~J Werbos.
\newblock 1990.
\newblock Backpropagation through time: what it does and how to do it.
\newblock {\em Proceedings of the IEEE}, 78(10):1550--1560.

\bibitem[\protect\citename{Weston \bgroup et al.\egroup
  }2015]{weston2015towards}
Jason Weston, Antoine Bordes, Sumit Chopra, Alexander~M Rush, Bart van
  Merri{\"e}nboer, Armand Joulin, and Tomas Mikolov.
\newblock 2015.
\newblock Towards ai-complete question answering: A set of prerequisite toy
  tasks.
\newblock {\em arXiv preprint arXiv:1502.05698}.

\bibitem[\protect\citename{Yu \bgroup et al.\egroup }2017]{yu2017deep}
Haonan Yu, Haichao Zhang, and Wei Xu.
\newblock 2017.
\newblock A deep compositional framework for human-like language acquisition in
  virtual environment.
\newblock {\em arXiv preprint arXiv:1703.09831}.

\bibitem[\protect\citename{Yuret}2016]{yuret2016knet}
Deniz Yuret.
\newblock 2016.
\newblock Knet: beginning deep learning with 100 lines of julia.
\newblock In {\em Machine Learning Systems Workshop at NIPS}.

\end{thebibliography}
\bibliographystyle{acl2012}

\end{document}